# Graphical Abstract

## Underwater object classification combining SAS and transferred optical-to-SAS Imagery

Avi Abu, Roee Diamant

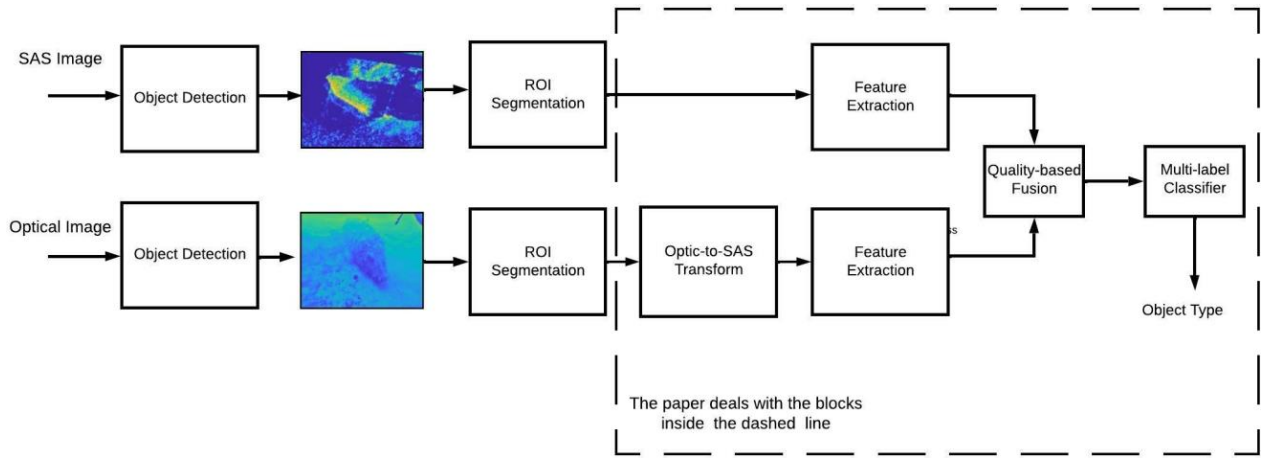

# Highlights

**Underwater object classification combining SAS and transferred optical-to-SAS Imagery**

Avi Abu, Roee Diamant

- A novel features set that uniquely characterizes the object's shape, and takes into account the object's highlight-shadow geometrical relations.

- A multi-label classification method for underwater objects using both optical and SAS images that doesn't require image calibration.

# Underwater object classification combining SAS and transferred optical-to-SAS Imagery

Avi Abu[a], Roee Diamant[a]

[a]The Hatter Department of Marine Technologies, Haifa University, 199 Aba Koushy Ave., Mount Carmel, Haifa, 3498838, Israel

**Abstract**

Combining synthetic aperture sonar (SAS) imagery with optical images for underwater object classification has the potential to overcome challenges such as water clarity, the stability of the optical image analysis platform, and strong reflections from the seabed for sonar-based classification. In this work, we propose this type of multi-modal combination to discriminate between man-made targets and objects such as rocks or litter. We offer a novel classification algorithm that overcomes the problem of intensity and object formation differences between the two modalities. To this end, we develop a novel set of geometrical shape descriptors that takes into account the geometrical relation between the object's shadow and highlight. Results from 7,052 pairs of SAS and optical images collected during several sea experiments show improved classification performance compared to the state-of-the-art for better discrimination between different types of underwater objects. For reproducability, we share our database.

*Keywords:* Feature extraction, shape descriptors, self-similarity, Fourier descriptor, region-based feature, contour-based features.

## 1. Introduction

Classification of underwater objects is a crucial step for underwater exploration tasks such as mine-hunting, pipeline monitoring, and seabed mapping. The general setup includes a surveying platform, and either a surface vessel or an autonomous underwater vehicle (AUV), which explores a given area to identify and classify targets on the seabed. Considering the non-homogeneous characteristics of the seabed with complex structures, such as sand ripples and rocks located beside patches of sand or sea-grass (e.g., Posidonia), the main challenge is to maintain low false alarm rates. The classification's aim is to enable the surveying vessel to act autonomously, so as to extend the mission times and allow for on-the-fly decision making. Furthermore, automatic classification combats the problems of operator fatigue and data deluge [1]. While most of the applications rely on the analysis of high-resolution synthetic aperture sonar (SAS), e.g., [2],[3], solutions also exist for the detection of optical images (e.g., [4]). As the current methods struggle to obtain low false alarm rates, one interesting concept is to combine the two modalities into a single sonar-optical classification system.

Combining optical and SAS data has the advantage of exploiting the sensory modalities for improved performance. While optical cameras offer a high resolution that can provide fine details of the object but require the surveying platform to be close to the object, SAS sensors have a higher range, but a lower resolution and provide fewer details. Specifically, the scenario considered here is of an AUV that surveys a given area, locks on a target, and then descends close to the seabed to provide a fine-resolution optical image of the target. In such a setup, a key enabling technology is to match the type of object detected within a sonar image and the type of target identified in the optical image.

The fusion of optical and sonar data has been applied before. In [5], a combination of a magnetic sensor, multi-beam sonar, and an optical camera is used for the task of mine hunting. Hover *et al.* [6] proposed a combination of optics and sonar to improve navigation applied to ship-hull inspection. In [7], three synchronized optical cameras were combined with the BlueView sonar system to form a photogrammetric system. Data fusion was based on matching 3D point clouds generated separately by each system. The solutions applied a 3D models by surface construction obtained by a triangulated mesh of each 3D voxel. However, this method involves an orientation phase to calibrate a set of acoustic and optical images, which increases the method's complexity and sensitivity to noise.

Multi-modal object matching can be categorized into two types: area-based and feature-based. The *area-based* technique uses a template of the desired object to match it with the inspected image, using metrics like cross-cumulative residual entropy [8], mutual information [9], and normal- ized cross-correlation coefficient [10]. Performance is highly dependent on the similarities between the inspected image and the template image. In matching sonar and optical images, this is a major challenge as the sensing techniques and resulting images are fundamentally different. Furthermore, area-based approaches are sensitive to scale, rotation, and translation differences. *Feature-based* methods search for similarities between the multi-modal images by matching features extracted from both modalities. Examples include geometrical features such as line, edge, object's contour properties [11], and point features [12]. Among these is the family of scale-invariant feature transform (SIFT) [13], which is robust to scale and rotation.

In this paper, we propose a multi-modal classification method for underwater objects in optical and SAS images. A pair of optical and sonar region of interest (ROI) images containing a target is presented by a preliminary detection process performed separately on each modality. We focus on the multi-hypothesis problem to examine whether these SAS and optical ROIs contain the same object type. To overcome intensity and object formation differences between the two modalities, we transform the object in the optical image to its equivalent model in SAS by the shape-from-shading method [14]. Since the object in the optical image has more details than in the SAS image, we choose to transform the optical to SAS to gain a more accurate acoustic model. During the transformation process, we take into account the geometry of the SAS system, the range between the object and the SAS, and the orientation of the object in the SAS image. We also take advantage of the object's shadow region, generated only in SAS imagery, which conveys important information about the object's type. Once the two modalities are merged, for the final classification of the object's type, we propose a new feature set that combines both the object's shadow and its highlight. Our main contributions are twofold:

1. A novel features set that uniquely characterizes the object's shape, and takes into account the object's highlight-shadow geometrical relations.
2. A multi-label classification method for underwater objects using both optical and SAS images that doesn't require image calibration.

We investigate performance using real SAS-optical image pairs that were obtained during multiple sea trials. The results show that our algorithm exceeds the state-of-the-art methods in terms of classification accuracy, and that the new features are useful for discriminating between object types in multi-modal images.

This paper is organized as follows. In Section 2, we survey the state- of-the-art of multi-modal image matching. In Section 3, the system model and main assumptions are presented. Section 4 contains the details of the proposed multi-modal object classification, while experimental results are presented in Section 5. Finally, concluding remarks are provided in Section 6.

## 2. Related Work

With the rapid development of satellites equipped with both SAR and optical systems, several techniques for matching the modalities have been proposed. In [15], a registration method is proposed to detect local feature points on the maximum moment map of the original image's phase congruency established through wavelet analysis. To overcome illumination and contrast variations, Li *et al.* [16] proposed the maximum index map (MIM), which is constructed by the log-Gabor convolution sequence. To increase the distinction between features, the authors used a distribution histogram technique to form a description vector. In [17], a SIFT-like method for optical-SAR image matching is proposed. To overcome intensity differences between the optical and SAR data, the authors used different gradient operators for the optical and SAR images, the multi-scale Sobel operator, and the two orthogonal 1-D infinite symmetric exponential filters (ISEF), respectively. In [18], a uniform nonlinear diffusion-based Harris (UND-Harris) feature extraction method is proposed to overcome noise in SAR and optical images. To handle nonlinear intensity variations in SAR and optical images, a phase congruency structural descriptor is proposed to establish optical-SAR image matching. In [19], geometrical features like edges and lines are identified. Compared to the SIFT descriptor, the authors reported improved performance in terms of the descriptors' matching rate.

While the above-mentioned methods can achieve satisfactory results in modalities like SAR and optical, none of them can be directly used for SAS and underwater optical data, due to the differences between the two modalities - specifically, the strong reflections from the seabed and the low resolution in SAS imagery [20], as well as the blurring due to highly variable lighting as a result of turbidity in optical images [21]. In addition, backscatter due to high water turbidity can affect the optical sensors' range and performance. Moreover, the same underwater scene produced by SAS and optical sensors is significantly different in terms of structural similarity and differences in intensity.

Recently, a few works focused on the problem of image matching in a sonar-optical underwater scene. In [22], a method for sonar-optical image matching was proposed. The matching process is based on the image's cross-correlation. First, the authors extract the image's spatial features. To handle differences in the object's scale, the authors utilized a Gaussian pyramid to build multi-scale images. In [23], a neural-network-based method for sonar-optical feature matching is proposed. Transfer learning was performed by adapting images over the VGG-19 network. Then, a SIFT-based feature detector was utilized by the nearest-neighbor algorithm to match between images. The work reported on undesirable acoustic targets, which decreased matching performance. Moroni *et al.* [24] proposed the matching of side-scan sonar and optical systems. First, features such as lines, circles, and ellipses are detected using the enhanced-ellipse-fitting method. Then, texture analysis is applied using the Gabor filters to discriminate between different seabed areas. The method requires accurate registration in order to apply sonar and optical data matching.

While the above surveyed sonar-optical image matching methods perform well in underwater environments, challenges are reported for image registration, and sensitivity to image quality. Considering this, we propose a novel underwater object classification algorithm - the SAS-optical image pair - which is less sensitive to image quality. Classification is made by fusing features of an object obtained from SAS and optical modalities. One way to fuse these features was proposed in [25] to distinguish between man-made and natural objects. This method did not evolve the optical-to-SAS transform and relies only on the geometrical features of the object. Here, to classify the type of man-made objects, we argue that the object's features are not sufficient, and the combination of the SAS and optical images should rely on more advanced characterizations. For example, the relation between the object and its shadow, which exists only in SAS imagery and has valuable information on the object's type. Similar to how it is done for multimodal optical images of different cameras or observation angles, this requires the merging of both modalities.

## 3. System Model

### 3.1. Main Assumptions

Our considered scenario is of an AUV surveying a given area in search of specific object types. Three classes of objects are considered: MANTA mines, $M$; cylinders, $C$; and natural objects, $N$. After detecting a possible target with its SAS system, the AUV navigates to take an optical image of the target. Given pairs of ROIs from SAS and optical images, each containing a single object, our goal is to decide whether the same type of object exists in both images. The classifier has four labels: $M, C, N, U$. The first three are determined if the object found in both SAS and optical images matches and is, in fact, one of the three target labels, while the fourth label, $U$, is declared if the object is not one of the three target types or if the sonar and optical images do not match. In doing so, our benefit is derived from the extra information obtained from analyzing both types of images. The decision plan is presented in Table 1. We consider the optic-to-SAS conversion and the segmentation as prior analysis. While the former may affect the quality of the transformed object from optic to SAS imagery, the latter related to the accuracy of the computed features of the object.

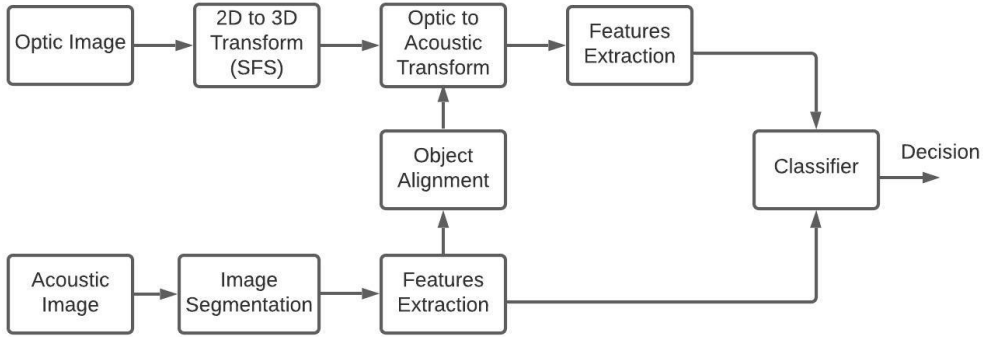

Figure 1: Block diagram of the proposed multi-modal object classification.

We make the following assumptions. First, we assume that a preliminary detection and segmentation process, performed separately on the SAS and optical images, has already detected a possible target and marked it over the image by a ROI. In our results, we used the methods in [26] and [20] for SAS detection and segmentation, respectively, and the methods in [4] and [27] for optical detection and segmentation, respectively. We further assume that both ROIs contain a single object. For practical reasons, the position where the SAS and optical images are taken is not constrained. That is, we avoid assuming that the AUV is stationed in the same position when taking the SAS and optical images, or that it is stationed at the same angle with respect to the object.

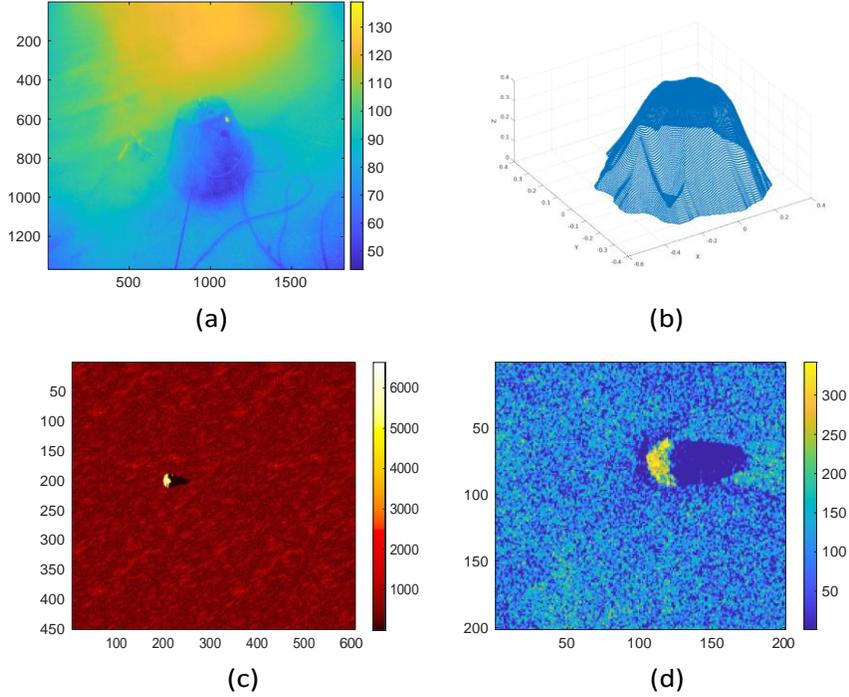

Figure 2: Examples of man-made (Manta-mine) objects in optical and SAS images. (a) An optical image obtained of Manta-mine. (b) 3d point clouds obtained from SFS [14]. (c) Synthetic SAS image we obtained from an optic-to-SAS transform. (d) A SAS image we obtained of Manta-mine.

Table 1: Ground-truth table of the classifier at the training phase

|  |  | Optical Image | | |
|---|---|---|---|---|
|  |  | M | C | N |
| SAS Image | M | M | U | U |
|  | C | U | C | U |
|  | N | U | U | N |

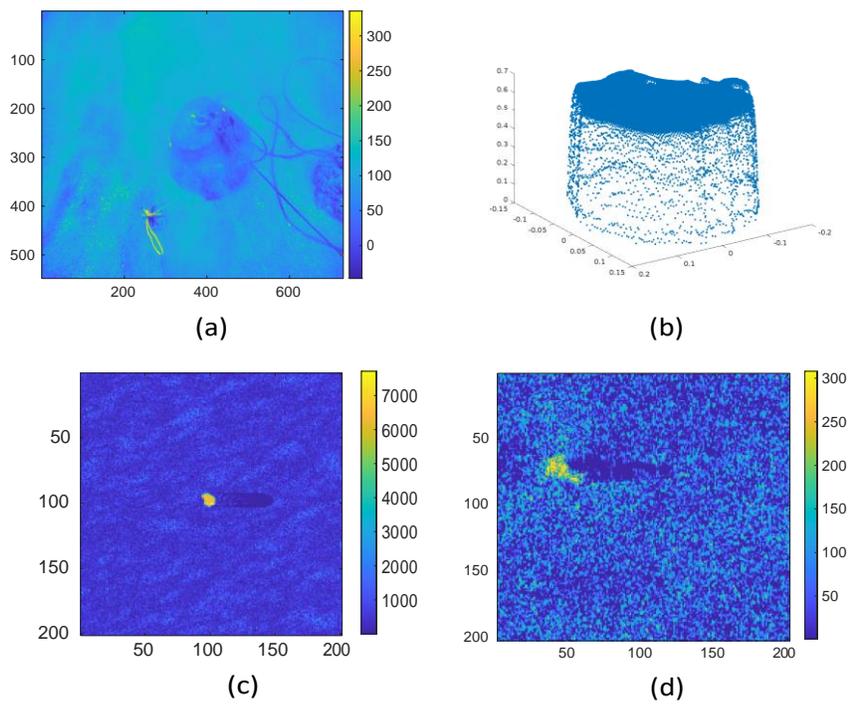

Figure 3: Examples of man-made (Manta-mine) object in optical image and cylinder in SAS image. (a) An optical image obtained of Manta-mine. (b) 3d point clouds obtained from SFS [14]. (c) Synthetic SAS image we obtained from an optic-to-SAS transform. (d) A SAS image we obtained of cylinder.

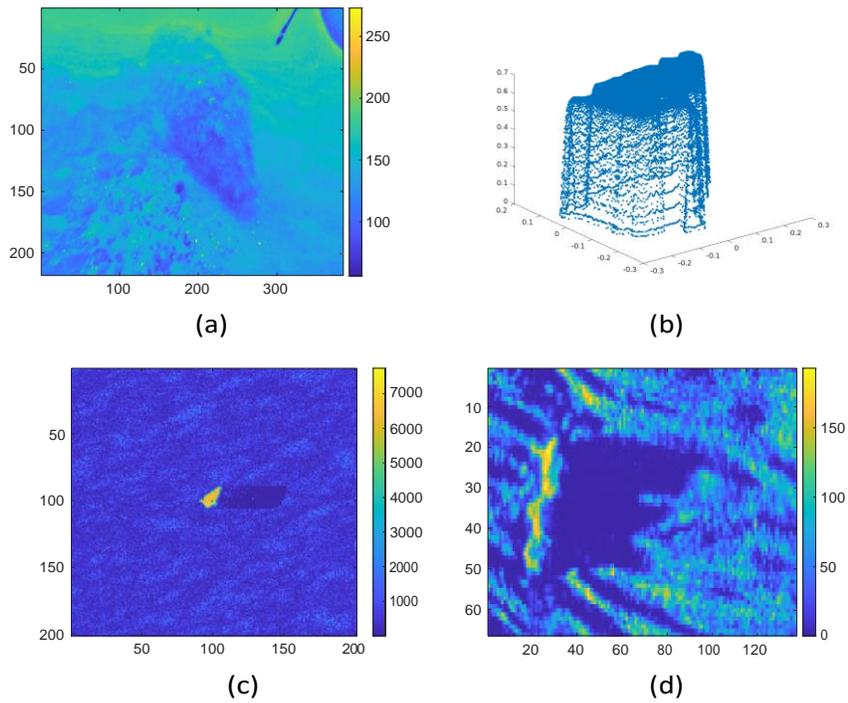

Figure 4: Examples of man-made (Cylinder) object in optical image and natural object in SAS image. (a) An optical image obtained of cylinder. (b) 3d point clouds obtained from SFS [14]. (c) Synthetic SAS image we obtained from an optic-to-SAS transform. (d) A SAS image we obtained of natural object.

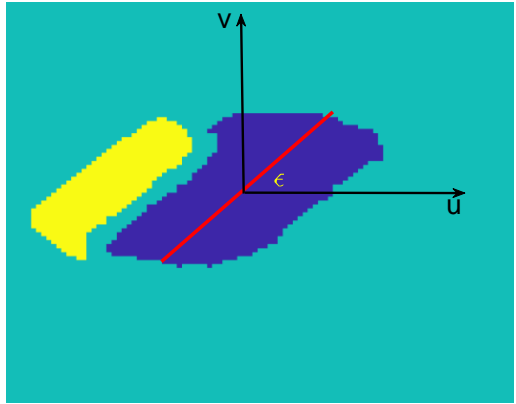

Figure 5: Measure of the orientation $\theta$ of the shadow zobe (blue region) for a cylindrical object. The red line corresponds to the major eigenvector. The orientation is given by the angle between the major eigenvector and the abscissa.

*3.2. Preliminaries*

In our method, we first convert the optical image to an SAS representation, and then perform classification using both SAS and SAS converted images. To enable a correct merging of the shadow properties between the SAS and this SAS-converted image, before conversion we first align the object in the optical image to match the orientation in the SAS image. This is because, in SAS representation, the object's orientation affects the geometrical characteristics of its shadow region. An examples of optic-to-SAS transformation with objects of the same category is given in Fig. 2. Fig. 3 and Fig. 4 show examples of a transformation process of misclassified objects.

The orientation $\theta$ of an object in a certain ROI is calculated by the angle of the eigenvector corresponding to the major eigenvalue of the object's covariance matrix. For a given ROI, the covariance matrix, $C_s$, is given by

$$C_s = \frac{1}{|\Delta|} \sum_{i \in \Delta} \tilde{v}_i \tilde{v}_i^T \tag{1}$$

where $\tilde{v}_i = (u_i - u_c, v_i - v_c)^T$, $(u_i, v_i), i \in \Delta$ are pixels within the shadow zone, and $(u_c, v_c)$ is the center of mass of the object in the ROI such that

$$v_c = \frac{1}{|\Delta|} \sum_{i \in \Delta} v_i, \tag{2}$$

and

$$c_b = \frac{1}{|\Delta|} \sum_{i \in \Delta} c_i.$$  (3)

The orientation is calculated by:

$$\theta = \arctan \frac{v_{c_b}(2)}{v_{c_b}(1)},$$  (4)

where $v_{c_b}$ is the eigenvector of $c_b$ related to the major eigenvalue, and $v_{c_b}(1)$ and $v_{c_b}(2)$ refer to the first and second value of the eigenvector, respectively. An orientation of a typical cylindrical object is given in Fig. 5. The yellow region refers to the object's echo, and the blue region refers to the shadow region. The red line corresponds to the major eigenvector.

## 4. Proposed Multi-modal Underwater Object Classification

### 4.1. Key Idea

A block diagram of our solution for the optic-SAS target classification task is illustrated in Fig. 1. Our key idea is to manage the different modalities by transforming the images into the same domain. Since SAS images are less detailed than optical images, a better conversion would be from optical to SAS. The scheme includes three steps: 1) optical to acoustic transformation, 2) object alignment, 3) features extraction, and 4) classification of object type. The first step involves a transformation of a 2D optical image into 3D, followed by conversion to an SAS representation. The second step includes segmentation of the images into highlight, shadow, and background regions. The third step includes feature extraction from both highlight and shadow regions. The final step includes classification of the feature outputs from both modalities.

### 4.2. Optic to Acoustic Transform

We convert the optical images to SAS and not vise versa, although it may cause information loss, since optical image contains more detailed information and higher resolution than SAS images that enable a good representation of the object after optic-to-SAS transform. To overcome intensity and structural differences between the SAS and optical modalities that affect the features' correspondence, an optic-to-SAS transform is performed prior to feature extraction. As referenced below, we note that parts of the following conversion steps are based on the work in [14]. The process includes four steps:

*4.2.1. Shape-from-Shading SFS*

First, the optical image is converted from 2D to 3D to yield information about the object's height, and thereby construct a shadow region. For 2D to 3D conversion, the image is segmented into background and highlight regions to identify the pixels that belong to the object. Then, the shape-from-shading (SFS) method [14] is executed to reconstruct the object's height by utilizing shading and illumination information from the 2D image. This approach showed promising results for underwater optical images [28]. The result is a 3D point cloud of the object that is centered to the origin of the converted image. An example of the operation of an SFS for a MANTA mine along with a real SAS image of the same object, are presented in Fig. 2.

*4.2.2. Object Alignment*

The posture of the object within the image holds information about the highlight reflection and the shape of the shadow zone in the SAS imagery. Since the objects in the optical and SAS images are likely to be sampled from different angles, an alignment of the object in the optic-to-SAS image onto the orientation of the object in the real SAS image is required. Since the highlight in SAS is more prominent than the shadow, we perform the alignment according to the orientation angle in the highlight region, $\theta_h$, for both the real SAS and optical-to-SAS images. We then rotate the 3D point cloud $\Theta$ of the object in the optical-to-SAS image, as obtained from SFS, by the difference between the highlight orientations of both images.

Let $\mathbf{f}^i \in \mathbb{R}^{3 \times 1}$, $i \in \Theta$ be a 3D point vector of the object given by SFS. The rotated vector $\mathbf{f}^i_r$ is given by

$$\mathbf{f}^i_r = \begin{bmatrix} \cos(\theta_r) & \sin(\theta_r) & 0 \\ -\sin(\theta_r) & \cos(\theta_r) & 0 \\ Z & 0 & 0 & 1 \end{bmatrix} \mathbf{f}^i, \tag{5}$$

where $\theta_r = (\theta_s - \theta_o)$, where $\theta_s$ and $\theta_o$ are the highlight's orientation of the SAS and optical images, respectively.

*4.2.3. Position of the Object in the Optic-to-SAS image*

The shape of the shadow and highlight in the SAS image is a function of the object's range and height, and the altitude of the AUV. As follows, the object's position in the optic-to-SAS image should be matched to the position of the object within the SAS image. To this end, we translate each 3D point vector by the center of the object ($□_□$, $□_□$) in the SAS image by the conversion,

$$\mathbf{f}'_□ = \mathbf{u}_□ - \mathbf{f}_□, \quad \forall □ \in \Theta, \tag{6}$$

where, $\mathbf{u}_□ = [□_□ \; □_□ \; 0]^□$.

*4.2.4. 3D Points Cloud to Synthetic Sonar Image*

To make the final conversion from an optical image to an SAS image, the visibility of each point from the 3D point cloud is estimated by the hidden-point-removal operator (HPR) [29]. We chose the HPR method due to its simplicity and efficiency to determine visible points out of a point cloud. The intensity of each point is thresholded to construct the shadow and highlight regions. Let $□_□ \in \mathrm{R}^{□_□ \times □_□}$ and $□_h \in \mathrm{R}^{□_□ \times □_□}$ be the binary representation of the object's shadow and highlight regions in the SAS image and the optic-to-SAS transformed image, respectively, such that the shadow map is given by

$$□_□(□, □) = \begin{cases} 1, & \text{if pixel with coordinates } (u,v) \text{ belongs to the shadow region} \\ 0, & \text{else} \end{cases}, \tag{7}$$

where ($□$, $□$) are the Cartesian coordinates of a pixel within the image. The highlight map, $□_h$ is defined in the same way as for the shadow.

*4.3. Proposed Features Set*

Recall that the input to the QDA classifier is the weighted feature set, $\mathbf{t}_□$, from the SAS and optical-to-SAS images. In this section, we describe in detail our proposed feature set. In our previous work [25], we proposed two new feature descriptors that relied on the smoothness of the object's contour to distinguish between man-made and natural objects. Here, we take a step forward and attempt to determine features that differentiate among the four target labels. The new features are based on the structure of the object's highlight and shadow regions and their geometrical relations, and aim to fit both the SAS and optical-to-SAS images.

The new features attempt to be scale and rotation-invariant to combat possible mismatches in the object's alignment process. To manage the different resolutions between the SAS and optical-to-SAS images, the new descriptors are normalized for scalability. We propose four features. The first, termed *Orientation Sum*, $\Gamma(\theta)$, describes the shadow along different angles of observations. The second feature executes a statistical analysis of $\Gamma(\theta)$, which can discriminate small changes in the orientation function in (9). The third, named *Highlight-Shadow Orientation* (HSO) reflects the geometrical relationship between the highlight and shadow regions. And the fourth, denoted as *Highlight Curvature* (HC), represents the curvature of the object's highlight.

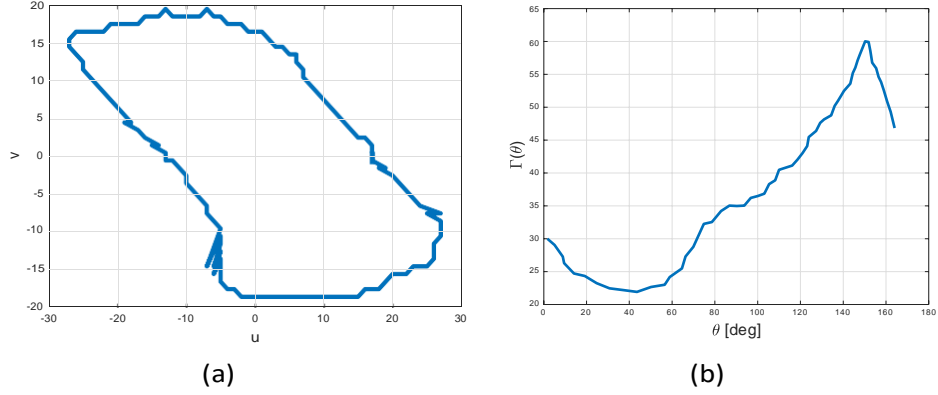

Figure 6: Orientation sum of a cylindrical object. (a) Contour of shadow region of a cylindrical object. (b) The orientation sum of the shadow region.

*4.3.1. Orientation Sum*

The orientation sum, $\Gamma(\theta)$, is defined as the discrete integration of $I_s(u, v)$ along a linear curve $L_\theta$ with orientation $\theta$ that crosses the center mass of the shadow region, $(u_c, v_c)$. Formally,

$$\Gamma(\theta) = \sum_{(u,v) \in L_\theta} I_s(u, v). \tag{8}$$

To manage the different scales, the orientation sum is normalized by

$$\overline{\Gamma}(\theta) = \frac{\Gamma(\theta)}{\max_\theta \Gamma(\theta)}. \tag{9}$$

We include in $\mathbf{t}_o$ two angles $\theta_{max}$ and $\theta_{min}$, for which $\bar{\Gamma}(\theta)$ gets its maximum and minimum values, respectively. That is,

$$\theta_{max} = \arg\max_\theta \bar{\Gamma}(\theta), \tag{10a}$$

$$\theta_{min} = \arg\min_\theta \bar{\Gamma}(\theta). \tag{10b}$$

Note that for symmetrical objects like a MANTA mine (see example in Fig. 2), $\theta_{max}$ and $\theta_{min}$ are invariant to the AUV posture relative to the object. An example of $\bar{\Gamma}(\theta)$ of a cylindrical object is presented in Fig. 6(b). The contour of the cylindrical object's shadow region is illustrated in Fig. 6(a). In this example, $\theta_{max}$ and $\theta_{min}$ are 151° and 43°, respectively.

*4.3.2. Discrete Wavelet Transform Features*

Although Fourier Descriptors (FD) have been widely used as geometrical features [30], these lack the capability to capture local variations in the object's shape. Instead, we propose a wavelet-based descriptor that can provide information with better resolution about local variations within the orientation sum $\bar{\Gamma}(\theta)$ in terms of both angle and frequency. More specifically, we use the discrete wavelet transform (DWT) of $\bar{\Gamma}(\theta)$ to quantify the orientation sum skewness $\gamma$. The skewness measures the asymmetry of the data distribution, and is independent of its mean. Formally, a skewness, $\gamma_s(s)$, along the scale axis $s$ is defined as [31]

$$\gamma_s(s) = \left[ \sum_\tau \left( \Omega_o(s, \tau) - \mu_\Omega \right)^3 \middle/ \left( \sum_\tau \left( \Omega_o(s, \tau) - \mu_\Omega \right)^2 \right)^{1.5} \right], \tag{11}$$

with

$$\mu_\Omega = \sum_s \sum_\tau \Omega_o(s, \tau). \tag{12}$$

The skewness along the translation axis $\tau$, $\gamma_\tau(\tau)$, is defined in the same way. The DWT feature of $\bar{\Gamma}(\theta)$, $\Omega_o$ is obtained by [32]

$$\Omega_o(s, \tau) = \sum_\theta \bar{\Gamma}(\theta) \psi_{s,\tau}(\theta), \tag{13}$$

where $s$ is the scale axis, $\tau$ is the translation axis. $\psi_{s,\tau}(\theta)$ is the Morlet wavelet, given by [33]

$$\psi_{s,\tau}(\theta) = \frac{1}{\sqrt{s\pi}} e^{-0.5\left(\frac{\theta-\tau}{s}\right)^2} \cos\left(2\pi f_0 \frac{\theta-\tau}{s}\right), \tag{14}$$

where $\sigma_b$ is the bandwidth parameter, and $\omega_0$ is the central wavelet frequency. The Morlet wavelet is used as the mother wavelet in this paper due to its capability to extract information related to subtle features in low and high-frequency components [34].

*4.3.3. Highlight-Shadow Orientation (HSO)*

We define the highlight-shadow orientation as the angle between the object's orientation in the highlight region, $\theta_h$, and the shadow region, $\theta_s$. The HSO feature reflects the geometric relationship between the shadow and highlight regions. The HSO is defined as

$$\text{HSO} = |\theta_s - \theta_h|, \qquad (15)$$

where $\theta_s$ and $\theta_h$ are the orientation of the object in the shadow and highlight regions, respectively. For a symmetrical object like a MANTA mine, the HSO is close to 90 degrees independently of the object and SAS pose. This is due to the orthogonality of the highlight and shadow orientations. For a cylindrical object, the HSO gets smaller values because its shadow and highlight regions have similar orientations. The HSOs of the MANTA mine and cylindrical objects, obtained from the SAS images in Fig. 7, are $82.26°$ and $56.18°$, respectively.

*4.3.4. Highlight Curvature (HC)*

Without loss of generality, consider that the object lies to the right of the AUV when the SAS and optical-to-SAS images are obtained. In this case, we extract a polynomial approximation of the left part of the highlight region as a geometrical feature that characterizes the complexity of the object's curvature. For example, a cylinder will have a simple structure relative to a natural object like a boulder. Let us define $\mathbf{q} = [q^1 \cdots q^i \cdots q^k]^\top$ as the coefficients vector of the $k$-th-order polynomial which, in terms of the L2 norm, best fits the left part of the highlight region. Formally,

$$\mathbf{q} = \underset{\mathbf{q} \in \mathbb{R}^{k \times 1}}{\arg\min} \sum_{i=1}^{N_h} \left( \mathbf{q}^\top \mathbf{u}_i - \frac{y_i}{x_i} \right)^2, \qquad (16)$$

where $\mathbf{p}_i = [x_i \; y_i]^\top$ is the coordinates of the $i$th point at the left part of the highlight region, and $\mathbf{u}_i = [1 \; x_i \ldots x_i^{k-1}]^\top$. $N_h$ is the number of pixels to the left of the highlight part. The left part of the highlight region is determined by taking all visible pixels (utilizing HPR [29]) of the object's contour when the viewpoint is from the left. The HC feature is defined by normalizing the coefficient vector **q** by its L2 norm for scale in-variance, such that

$$\hat{q} = \frac{q}{\|q\|_2}. \qquad (17)$$

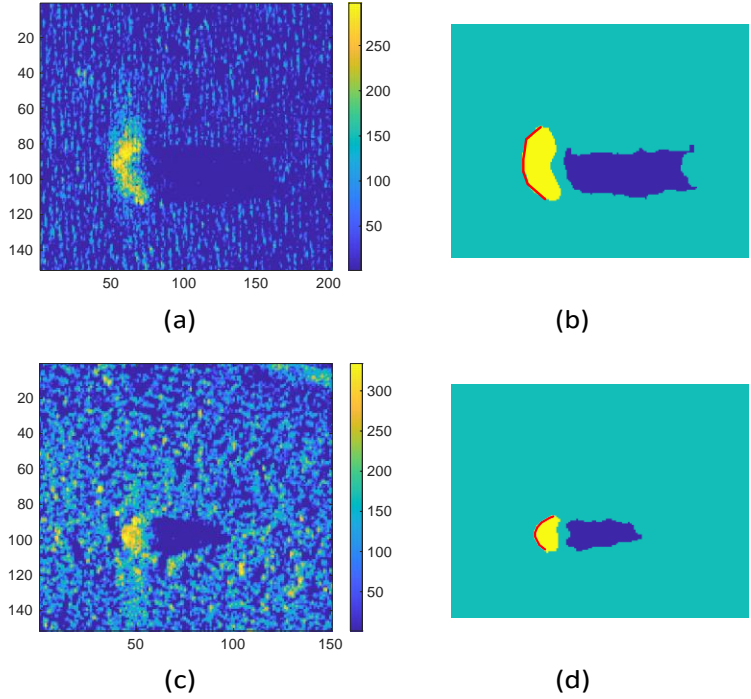

Figure 7: Difference between the curvature of the object's highlight. (a) SAS image we obtained of a cylindrical object. (b) Segmentation map of the SAS image. (c) SAS image we obtained of a Manta-mine object. (d) Segmentation map of the SAS image. The red line is the curvature of the left part of the object's highlight.

Fig. 7 shows an example of a cylinder and MANTA mine obtained from SAS imagery. The red line indicates the left part of the object's highlight. The HC of the cylinder is: $[0.98\ 0.18\ 0.01\ 0.0004]^\top$, and the HC of the MANTA mine is: $[0.78\ 0.6\ 0.16\ 0.02]^\top$. This implies that the MANTA mine's curvature is more complex than that of a cylinder.

*4.4. Multi-Label Classifier*

A decision on the existence of a type of man-made object in an optic-SAS image pair is made by a multi-label classifier that is trained over the geometrical features of both SAS and converted optic-to-SAS. For this task, we chose the quadratic discriminant analysis (QDA) classifier [35], due to its advantage when the training set size is small. To handle cases of a low-quality image that affects the accuracy of the feature vector, we propose a quality-based merging technique of the feature vectors obtained from the SAS and the converted optic-to-SAS images. The process begins with a segmentation of the SAS and optical-to-SAS images. In our results, we use our previous work [36], but other segmentation solutions are possible. The classification steps are described below.

*4.4.1. Quadratic Discriminant Analysis (QDA)*

We first explain how we adapt the QDA technique to our problem. Recall that four labels are handled by the classifier: MANTA mines **M**, cylinders **C**, natural objects **N**, and **U**. The classifier is fed with a weighted feature vector $\mathbf{t}_\square$ obtained from the SAS and converted optic-to-SAS feature vectors $\mathbf{t}_\square$ and $\mathbf{t}_\square \in \mathrm{R}^{1 \times \square 1}$. Since QDA handles only one input set, we merge the two images by

$$\mathbf{t}_\square = \bar{\square}_\square \mathbf{t}_\square + \bar{\square}_\square \mathbf{t}_\square, \tag{18}$$

where $\bar{\square}_\square = \frac{\square_\square}{\square_\square + \square_\square}$ and $\bar{\square}_\square = \frac{\square_\square}{\square_\square + \square_\square}$, and coefficients $\square_\square$ and $\square_\square$ are scalar parameters that quantify the quality of the SAS and optical-to-SAS images, respectively. Our proposal for how to determine these quality parameters is given in the subsection below. We assume that the probability density function (PDF) of both feature vectors $\mathbf{t}_\square$ and $\mathbf{t}_\square$ is Gaussian. That is,

$$\square(\mathbf{t}) = \frac{1}{\square_\square (2\square)^{\square/2}} |\Sigma^\square|^{-1/2} \exp\left(-\frac{1}{2}(\mathbf{t} - \bar{\mathbf{t}}_\square)^\square (\Sigma^\square)^{-1}(\mathbf{t} - \bar{\mathbf{t}}_\square)\right), \quad \square = M, C, N, U, \tag{19}$$

where $\bar{\mathbf{t}}_\square^\square$ and $\Sigma_\square^\square$ are the mean vector and the covariance matrix of the $l\square h$ class, respectively, obtained from the weighted feature vector, and are given by

$$\bar{\mathbf{t}}_\square^\square = \bar{\square}_\square \bar{\mathbf{t}}_\square^\square + \bar{\square}_\square \bar{\mathbf{t}}_\square^\square, \tag{20}$$

and

$$\Sigma_\square^\square = \bar{\square}_\square^2 \Sigma_\square^\square + \bar{\square}_\square^2 \Sigma_\square^\square. \tag{21}$$

---

[1]We note that, in principle, any feature set is possible. In this work, part of our contribution is the proposal of a new feature set which, based on our dataset, produces good classification results. A description of this new feature set is presented in Section 4.3 below.

Table 2: L2 norms of estimated covariance matrix $\hat{\Sigma}^\square$ and estimated cross-covariance term $\hat{\square}\ \overline{\square}_\square \overline{\square}_\square (\mathbf{t}_\square - \overline{\mathbf{t}}_\square)(\mathbf{t}_\square - \overline{\mathbf{t}})^\square$ for the C, N, M, and U classes. All quantities are normalize )

| Object Type | $\hat{\Sigma}^\square_\square$ | $\overline{\square}_\square \overline{\square}_\square (\mathbf{t}^\square - \overline{\mathbf{t}}^\square)(\mathbf{t}^\square - \overline{\mathbf{t}}^\square)$ |
|---|---|---|
| M | 1.4 | 0.2 |
| C | 1.7 | 0.11 |
| N | 2.2 | 0.14 |
| U | 1.43 | 0.16 |

In (21), we neglect the cross-covariance terms. To evaluate this hypothesis, Table 2 shows the L2 norms of the estimated covariance matrix $\hat{\Sigma}^\square_\square$ and the estimated cross-covariance matrix $\hat{\square}\ \overline{\square}_\square \overline{\square}_\square (\mathbf{t}_\square - \overline{\mathbf{t}}_\square)(\mathbf{t}_\square - \overline{\mathbf{t}})^\square$ for each class of object. Results are obtained with $\grave{\square}^\square_\square = 0.1$, $\square^\square_\square = 0.25$, and $\square^h_\square = 0.3$. Here, $\overline{\mathbf{t}}_\square$ and $\overline{\mathbf{t}}_\square$ are the mean vectors of the $l$-$h$ label obtained from the SAS and converted optic-to-SAS feature vectors, and similarly, $\Sigma^\square_\square$ and $\Sigma^\square_\square$ are the covariance matrices. We note that these mean and covariance matrices are calculated from a training data set that can be real SAS images (as in our case) or modeled simulations using, for example, the simulator in [37].

The QDA classifier assigns the feature vector $\mathbf{t}_\square$ to a label $\square$ if

$$\log \frac{\square_\square(\mathbf{t}_\square)}{\square_\square(\mathbf{t}_\square)} > 0, \quad \forall \square \neq \square, \quad (\square, \square) \in M, C, N, U, \tag{22}$$

*4.4.2. Determining the Quality of the Images*

To determine the ratio parameters in (18), we evaluate the quality of the SAS and optical-to-SAS images. Our intuition to quantify the quality of an image, is that an image with clearly separated background and highlight regions produces a better classification. Thus, inspired by [38], we propose to determine the quality of the SAS and optical images by a quality index $\square$,

$$\square = \frac{(\square - \square_{\square\square})^2}{\square^2_{\square\square} + \square^2_{\square\square}}, \tag{23}$$

where $\square_{\square\square}$ and $\square_{\square\square}$ are the mean of the images' pixels belonging to the background and the echo labels, respectively. The labels are obtained from the preliminary segmentation process. The variances of the background and echo are denoted by $\sigma_n^2$ and $\sigma_e^2$. Due to different resolution and statistical models of the noise in SAS and optical modalities, we cannot directly apply $\sigma_n$ in the merging process in (18). Instead, we propose a transfer function $\Psi$ with parameter set $\{\alpha_o^l, \alpha_o^h, \alpha_o^h, \beta_o^l, \beta_o^h\}$ for the SAS image and $\{\alpha_s^l, \alpha_s^h, \alpha_s^h, \beta_s^l, \beta_s^h\}$ for the optical image. These parameters are fine-tuned by trial-and-error using the data in the training set. The transfer function is then defined by

$$\bar{\sigma}_s = \Psi_s(\sigma_s) = \begin{cases} \alpha_s^l, & \sigma_s < \beta_s^l \\ \alpha_s^o, & \beta_s^l \le \sigma_s < \beta_s^h \\ \alpha_s^h, & \sigma_s \ge \beta_s^h \end{cases}, \quad (24)$$

where $\sigma_s$ is the quality index of the SAS image. The quality function $\Psi_o(\sigma_o)$ for the optic image is defined similarly for the quality index $\sigma_o$.

## 5. Experimental Results

We compare our algorithm with three state-of-the-art methods, i.e., the SSC [39], IDSC [40], and HF [41], all represent different approaches to object matching and have been previously used for sonar image processing. The implementation of the benchmark is obtained from the authors' website. For the discrete wavelet features, the bandwidth and central frequency $f_b$ and $f_c$ are set to 1.5 and 1, respectively. The polynomial order is set to $L = 3$. The parameters of the optic transfer function $\Psi_o$ are set to: $\alpha_o^l = 0.3$, $\alpha_o^o = 3$, $\alpha_o^h = 0.1$, $\beta_o^l = 0.25$, and $\beta_o^h = 0.3$. The parameters of the acoustic transfer function $\Psi_s$ are set to be: $\alpha_s^l = 0.3$, $\alpha_s^o = 3$, $\alpha_s^h = 0.4$, $\beta_s^l = 0.6$, and $\beta_s^h = 0.65$. These parameters are set by trial-and-error using the data in the training set. The classifier is trained by 70% of the image pairs, which are randomly uniformly chosen. Experimental results are obtained by running 50 independent Monte-Carlo trials, and then averaging the results. The multi-label classification results are obtained by the "one-vs.-rest" technique. That is, for each class, we consider a binary classification problem by considering the rest classes as a "negative" class. Then, we average the classification results obtained for each class to extract the multi-class result. Formally, the object is classified as M if $\Gamma_{c,c'}(\mathbf{t}_c) > 0$ for $c \in$ M and $c' \in$ C, N, U, and analogously for C, N, and U. Then, the final result is obtained by averaging the above four classification results.

Table 3: Data Set collection details

| Object Type | Opportunities | Sensor Type |
|---|---|---|
| Cylinder | 31 | SAS |
|  | 35 | optic |
| Manta | 25 | SAS |
|  | 11 | optic |
| Natural objects | 30 | SAS |
|  | 36 | optic |

*5.1. Dataset*

We obtained our sonar dataset using a Kraken-made, two-sided synthetic aperture sonar (SAS), mounted on our A18 5.5m Eca Robotics Inc. AUV [42]. The sonar images were collected opportunistically by programming the AUV with a mission to explore different areas across the Israeli Mediterranean coastline. In the first experiment, the area was roughly 2 miles west of the Caesarea Port and the AUV was at a depth of 25 m over a rocky seabed. In the second experiment, the area was roughly 2 miles west of the Haifa Port at a depth of 20 m. The third experiment was conducted about 10 km west of Northern Israel at a water depth of 1,000 m. After recovering the vehicle, we observed the images manually and identified "interesting objects" - both man-made and natural. In the first two experiments, we also sent scuba divers to photograph the identified objects as well as the underwater surroundings using hand-held underwater cameras. The optical cameras had a twelve mega-pixels sensor, and produced $4000 \times 3000$ pixel images. A total of 86 SAS images and 82 optical images were collected. The database contains SAS and optical images. Each image in each modality is labled by one of the three labels: cylinders, manta-mine, and natural objects. In our analysis, we take into account all possible pairs between the optical and SAS images to obtain a total number of image pairs (SAS-optic) of 7,052. We divided our dataset into object types. The numbers are reported in Table 3. Examples of such types are given in Fig. 8 for Manta-mine, cylinder and boulder. For reproducibility, we share our database in [43].

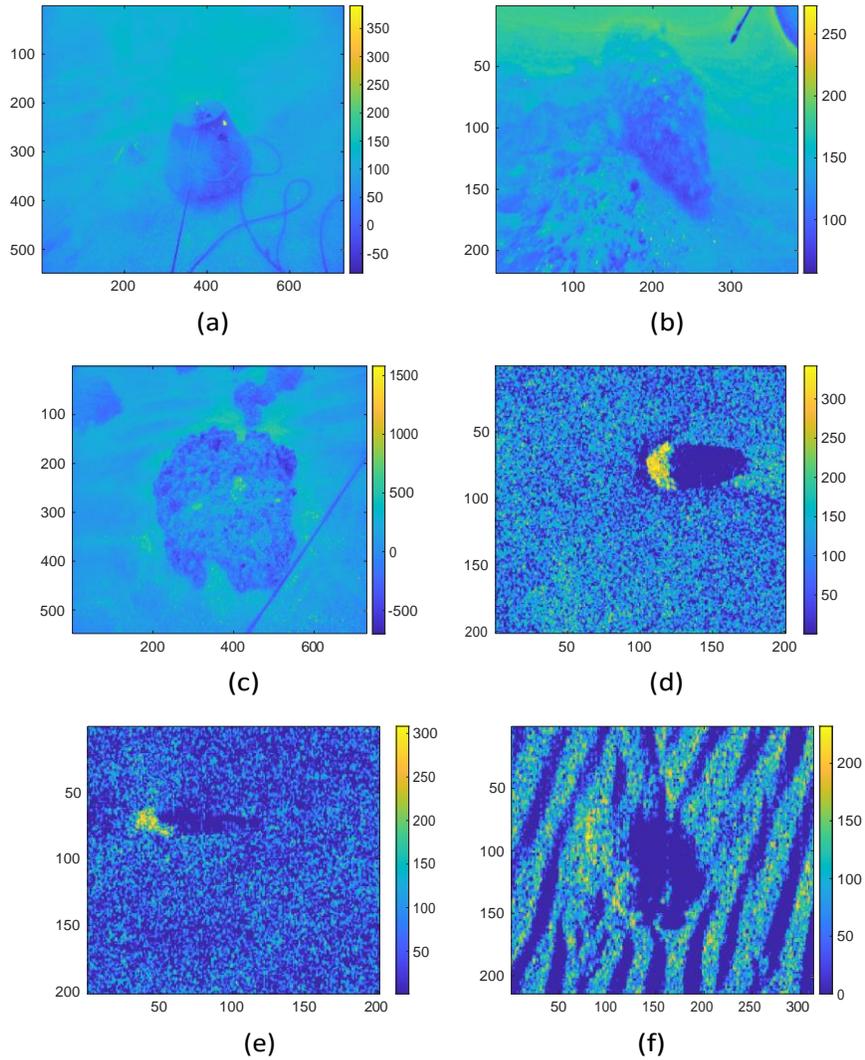

Figure 8: Examples of optical and SAS images contain manta, cylinder, and boulder type objects. Optical images: (a) Manta. (b) Cylinder. (c) Boulder. SAS images: (d) Manta. (e) Cylinder. (f) Boulder.

*5.2. Comparison with the Benchmark*

In Fig. 9(a), we compare the performance of our approach with the three benchmarks in terms of the receiver-operating-characteristic (ROC) curves for a quadratic discriminant analysis (QDA) classifier. A similar comparison is shown in Fig. 9(b) and Fig. 9(c) for two more classifiers: the linear discriminant analysis (LDA) and the support vector machine (SVM), respectively. Results for all methods seem better for the QDA classifier compared to those of the LDA classifier. This is because the classes have different covariance matrices and cannot be modeled by a single one, as the LDA assumes. In this case, the LDA suffers from a large bias. We also observe that the results of the HF are the closest to our proposed approach. This is due to its robustness to local deformations that occur during the segmentation of the object's shape. Table 4 introduces the benchmark's confusion matrices and the proposed method for the QDA classifier. We observe that while the results for our approach and HF have similar probabilities of identifying objects with the labels M and U, our proposed method outperfoms HF when it comes to identifying objects with the label N (above 0.9). Furthermore, for objects with the N label, the HF false alarm is about 4 times greater than that of our proposed method. This is because our method takes into account the geometric relationship between the shadow and highlight regions, rather than considering only the shape structure of the object's contour, and in complex structures like N, it helps to discriminate between objects.

*5.3. Polynomial Order Analysis*

The highlight curvature HC feature characterizes the complexity of the left part of the object's highlight region. Fig. 10 introduces the ROC for different values of $\square$. We observe that the case $\square = 1$ achieves the lowest performance, while no significant differences are shown between the results obtained with $\square = 2$ and $\square = 3$. We thus conclude that linear approximation is not well suited to concave-like structures such as boulders and MANTA mines. Also, utilizing polynomial order above three decreases classification performance.

*5.4. Analytic Wavelet Impact on the Classification Results*

In this section, we analyze the impact of the chosen analytic wavelet function on the classifier's performance. We test two wavelet functions: Morlet and Morse. The first is a sinusoidal-based function, while the latter is exponential-based. ROC for both functions are introduced in Fig. 11. We observe that, when using Morlet, better performance is obtained than when using Morse. This is mainly due to Morlet's capability to capture fine discrimination in low and high-frequency components.

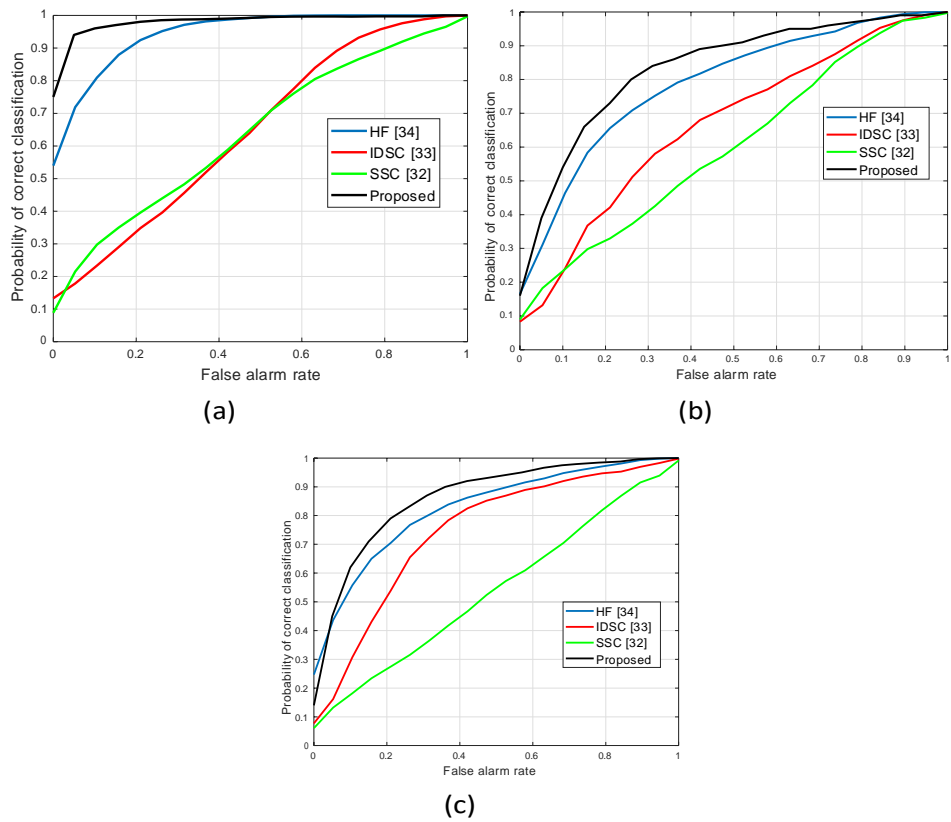

Figure 9: Classification performance for the proposed method and the benchmark with different classifiers. (a) Quadratic discriminant analysis (QDA). (b) Linear discriminant analysis (LDA). (c) Support vector machine (SVM).

Table 4: Confusion matrices of the benchmarks and the proposed method obtained with quadratic discriminant analysis (QDA) classifier

| **SSC** | Decision Label | | | |
|---|---|---|---|---|
| Ground Truth | M | C | N | U |
| M | 0.8 | 0.15 | 0 | 0.05 |
| C | 0.48 | 0.41 | 0 | 0.01 |
| N | 0.49 | 0.27 | 0.18 | 0.06 |
| U | 0.59 | 0.25 | 0.09 | 0.07 |

| **IDSC** | Decision Label | | | |
|---|---|---|---|---|
| Ground Truth | M | C | N | U |
| M | 0.11 | 0.35 | 0 | 0.54 |
| C | 0.11 | 0.34 | 0.16 | 0.39 |
| N | 0.07 | 0.29 | 0.52 | 0.12 |
| U | 0.02 | 0.42 | 0.08 | 0.48 |

| **HF** | Decision Label | | | |
|---|---|---|---|---|
| Ground Truth | M | C | N | U |
| M | 0.95 | 0.02 | 0 | 0.03 |
| C | 0.04 | 0.72 | 0.08 | 0.16 |
| N | 0.02 | 0.02 | 0.75 | 0.21 |
| U | 0.02 | 0.11 | 0.11 | 0.76 |

| **Proposed** | Decision Label | | | |
|---|---|---|---|---|
| Ground Truth | M | C | N | U |
| M | **0.96** | 0.01 | 0.03 | 0 |
| C | 0 | **0.83** | 0.11 | 0.06 |
| N | 0 | 0.04 | **0.94** | 0.02 |
| U | 0.07 | 0 | 0.14 | **0.79** |

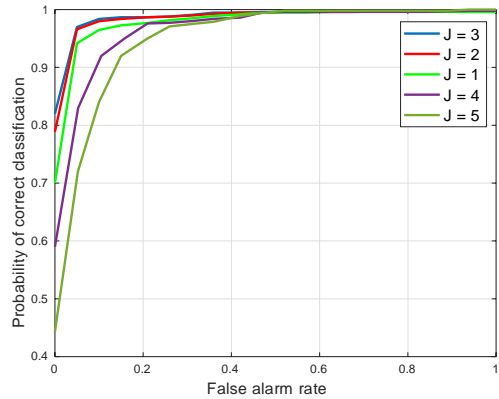

Figure 10: Comparison of the classification performance for different values of polynomial order ☐. Best results are obtained with ☐ = 3. As expected, for linear approximation case (☐ = 1), the classification results are degraded.

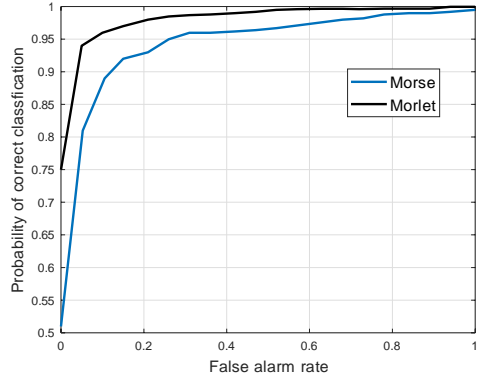

Figure 11: Comparison of the classification performance with different wavelet functions.

*5.5. SAS and Optic Feature Set Merging*

Next, in Fig. 12(a) we compare the performance of our proposed quality-based merging method with an average merging for the SAS and optical feature set. We observe that the former outperforms the averaging approach. This implies that the quality of the SAS and optical images holds valuable information for classification. Fig. 12(b) shows the parameter sets for the transfer function $\Psi_\square$. In this analysis, the impact of transfer function parameters on classification results is investigated. We chose to change the optic parameters, but similar results are obtained when changing the acoutic ones. Here, set A is defined for $\alpha^\square_\square = 0.2$, $\alpha^\square_\square = 0.4$, and $\alpha^h_\square = 0.6$. Set $\square$ is defined for: $\alpha^\square_\square = 0.25$, $\alpha^\square_\square = 0.5$, and $\alpha^h_\square = 0.75$, and set $\square$ is defined for: $\alpha^\square_\square = 0.1$, $\alpha^\square_\square = 0.25$, and $\alpha^h_\square = 0.3$. We observe that the choice of parameters does, in fact, impact the performance, with set $\square$ achieving the best performance. Table 5 introduces the confusion matrices obtained with quality-based and averaging methods. To analyze the sensitivity of the alpha-parameters, we generate 500 random (Gaussian distributed with STD equals 0.4) alpha-parameters set centralized in the values obtained by trial and error: $\alpha^\square_\square = 0.1$, $\alpha^\square_\square = 0.25$, and $\alpha^h_\square = 0.3$. For each set, we have calculated the average of the true positives of all classes obtained by averaging the diagonal of the confusion matrix. The normalized histogram of the results is introduced in Fig. 13. The average of the true positives obtained at the trial-and-error point is 0.88, as can be seen from the results. By the sharpness of the histogram we conclude that the proposed approach is not sensitive for the selection of the alpha-parameters.

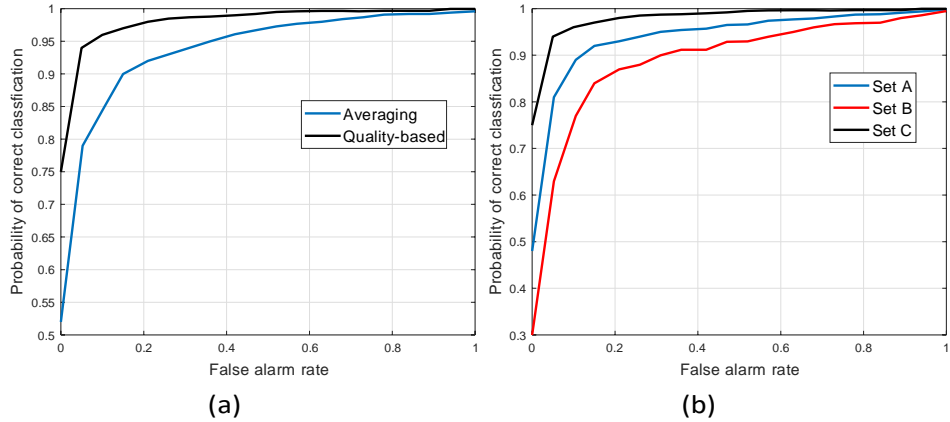

Figure 12: Feature set merging analysis. (a) Comparison of the classification performance with quality-based and averaging merging approaches. (b) ROC curves for different transfer function parameters.

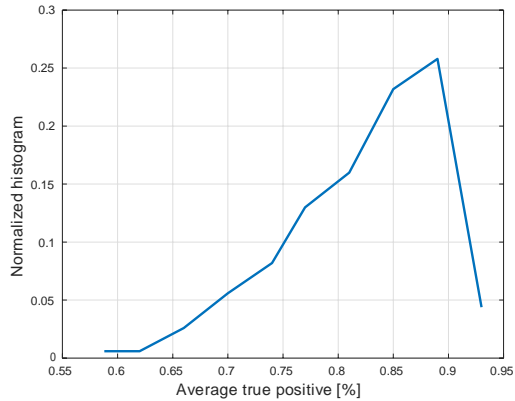

Figure 13: Histogram of average true positive of all classes obtained by 500 random sets of $\{\alpha_i^b, \alpha_i^p, \alpha_i^h\}$.

## 5.6. Merging Data vs. Single Sensor Data

Finally, we explore the benefit of the SAS and optical data combination. The classifier's performance using SAS data only, optical data only, and their combination are shown in Fig. 14. In this analysis, all features are taken from the SAS and transformed-optic-to-SAS images. In optic-only classification, we set $\Box_\Box = 1$ and $\Box_\Box = 0$, namely, features are taken from the transformed optic-to-SAS images. We observe a rather similar performance for the SAS-only and optical-only approaches, which reflects the equal information provided by both modalities. Yet, when combining the data, an observable improvement results, suggesting that mutual information exists in both images, and thus a diversity gain. The confusion matrices for SAS only, projected optic-to-SAS only, and quality-based SAS-optic merging are included in Table 6.

Table 5: Confusion matrices obtained with averaging and quality-based SAS-optic merging techniques

| Averaging | Decision Label | | | |
|---|---|---|---|---|
| Ground Truth | M | C | N | U |
| M | 0.8 | 0.02 | 0.03 | 0.15 |
| C | 0 | 0.71 | 0.13 | 0.16 |
| N | 0 | 0 | **0.97** | 0.03 |
| U | 0.08 | 0.07 | 0.12 | 0.73 |

| Quality-based | Decision Label | | | |
|---|---|---|---|---|
| Ground Truth | M | C | N | U |
| M | **0.96** | 0.01 | 0.03 | 0 |
| C | 0 | **0.83** | 0.11 | 0.06 |
| N | 0 | 0.04 | 0.94 | 0.02 |
| U | 0.07 | 0 | 0.14 | **0.79** |

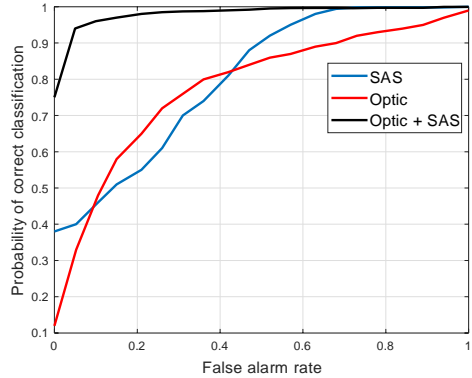

Figure 14: Classification results obtained from SAS data only, optical data only, and quality-based SAS-optical data merging.

The mean and standard deviation of the proposed features are included in Table 7. $\Box\Box_1$ and $\Box\Box_2$ are the second and third components of $\Box\Box$. As expected, due to the symmetrical property of the manta, $\Box_{\Box\Box\Box}$ and $\Box_\Box$ are approximately equal, while cylinder has its $\mu_{\square\square}$ and $\mu_{\square\square}$ approximately, around zero, and 90 degrees, respectively. While $\mu_{\square\square}$ is not a good descriptor, $\mu_{\square\square}$ is useful for M and C discrimination. $\square\square_2$ represents the non-linear term in the L2 approximation, and can efficiently discriminate between M and C, while $\square\square_1$ is less effective since it stands for the linear term in the L2 approximation.

Table 6: Confusion matrices obtained from SAS data only, projected optic-to-SAS data only, and quality-based SAS-optical data merging

| SAS only | Decision Label | | | |
|---|---|---|---|---|
| Ground Truth | M | C | N | U |
| M | 0.78 | 0.03 | 0.02 | 0.17 |
| C | 0 | 0.7 | 0.1 | 0.2 |
| N | 0 | 0 | 0.82 | 0.18 |
| U | 0.05 | 0.05 | 0.14 | 0.76 |

| Optic Only | Decision Label | | | |
|---|---|---|---|---|
| Ground Truth | M | C | N | U |
| M | 0.78 | 0.04 | 0.02 | 0.16 |
| C | 0 | 0.65 | 0.16 | 0.19 |
| N | 0 | 0 | 0.78 | 0.12 |
| U | 0.08 | 0.08 | 0.12 | 0.72 |

| Proposed | Decision Label | | | |
|---|---|---|---|---|
| Ground Truth | M | C | N | U |
| M | **0.96** | 0.01 | 0.03 | 0 |
| C | 0 | **0.83** | 0.11 | 0.06 |
| N | 0 | 0.04 | **0.94** | 0.02 |
| U | 0.07 | 0 | 0.14 | **0.79** |

Table 7: Mean and standard deviation (in brackets) of the proposed features for the C, N, M, and U classes

| Feature | $\square_{\square\square\square}$ | $\square_{\square\square\square}$ | $\square$ | HSO | $\square\square_1$ | $\square\square_2$ |
|---|---|---|---|---|---|---|
| M | 78 (12) | 92 (18) | 0.15 (0.1) | 83 (12) | 0.12 (0.4) | 0.15 (0.1) |
| C | 4 (2) | 75 (12) | 0.36 (0.12) | 65 (31) | 0.45 (0.52) | -0.03 (0.14) |
| N | 85 (32) | 65 (43) | -0.14 (0.16) | 70 (40) | -0.3 (0.6) | 0.27 (0.3) |
| U | 48 (21) | 81 (32) | 0.08 (0.22) | 61 (32) | 0.02 (0.1) | 0.1 (0.28) |

## 6. Conclusion

In this paper, we proposed a novel classification method for underwater objects combining optical and SAS imagery. To overcome intensity and object formation differences between the two modalities, we transform the optical image into an equivalent SAS image. Then, classification is performed by fusing a set of descriptors that quantify the geometrical relationship between the object's shadow and highlight. Experimental results, obtained from a large dataset of real SAS and optical image pairs, demonstrate accurate classification performance compared to the state-of-the-art. In addition, a quantitative evaluation demonstrates the effectiveness of our proposed fusion method. The proposed method relies on the existence of highlight and shadow regions in SAS imagery. While highlight is more prominent, shadow may not always exist due to imperfect ownship compensation. The accuracy of the object's feature relies on the quality of the optic-to-SAS transform as well as the segmentation process, which, in this paper, considered as preliminary processes. Future work is recommended to combine the feature descriptors to verify that the same object exists in both SAS and optical images.